\documentclass[sn-mathphys-num]{sn-jnl}

\usepackage{graphicx}%
\usepackage{multirow}%
\usepackage{amsmath,amssymb,amsfonts}%
\usepackage{amsthm}%
\usepackage{mathrsfs}%
\usepackage[title]{appendix}%
\usepackage{xcolor}%
\usepackage{textcomp}%
\usepackage{manyfoot}%
\usepackage{pifont}
\usepackage{url}
\usepackage{verbatim}
\usepackage{textcomp}
\usepackage{stfloats}
\usepackage{algorithm}
\usepackage{array}
\usepackage{color}
\usepackage{ulem}
\usepackage[caption=false,font=normalsize,labelfont=sf,textfont=sf]{subfig}
\usepackage{booktabs}%
\usepackage{algorithm}%
\usepackage{algorithmicx}%
\usepackage{algpseudocode}%
\usepackage{listings}%

\theoremstyle{thmstyleone}%
\theoremstyle{thmstyletwo}%

\theoremstyle{thmstylethree}%

\newcommand{\eg}{\emph{e.g., }}

\begin{document}

\title[Article Title]{3D Hand Mesh Recovery from Monocular RGB in Camera Space}


\author[1]{\fnm{Haonan} \sur{Li}}\email{lihaonan.scut@gmail.com}

\author*[1]{\fnm{Patrick P. K.} \sur{Chan}}\email{patrickchan@ieee.org}

\author*[1]{\fnm{Yitong} \sur{Zhou}}\email{zhouyitong@scut.edu.cn}

\affil[1]{\orgdiv{Shien-Ming Wu School of Intelligent Engineering}, \orgname{South China University of Technology}, \orgaddress{\street{777 Xingye Road}, \city{Guangzhou}, \postcode{510000}, \state{Guangdong}, \country{China}}}


\abstract{With the rapid advancement of technologies such as virtual reality, augmented reality, and gesture control, users expect interactions with computer interfaces to be more natural and intuitive. Existing visual algorithms often struggle to accomplish advanced human-computer interaction tasks, necessitating accurate and reliable absolute spatial prediction methods. Moreover, dealing with complex scenes and occlusions in monocular images poses entirely new challenges. This study proposes a network model that performs parallel processing of root-relative grids and root recovery tasks. The model enables the recovery of 3D hand meshes in camera space from monocular RGB images. To facilitate end-to-end training, we utilize an implicit learning approach for 2D heatmaps, enhancing the compatibility of 2D cues across different subtasks. Incorporate the Inception concept into spectral graph convolutional network to explore relative mesh of root, and integrate it with the locally detailed and globally attentive method designed for root recovery exploration. This approach improves the model's predictive performance in complex environments and self-occluded scenes. Through evaluation on the large-scale hand dataset FreiHAND, we have demonstrated that our proposed model is comparable with state-of-the-art models. This study contributes to the advancement of techniques for accurate and reliable absolute spatial prediction in various human-computer interaction applications.}

\keywords{Hand shape estimation, 3D computer vision, Spectral graph CNN}



\maketitle

\section{Introduction}\label{sec1}

Monocular 3D mesh recovery aims to extract 3D locations of mesh vertices from a single image. A precise 3D mesh enhances the authenticity of virtual reality (VR) and augmented reality (AR) technologies \cite{Kong2022IdentityAwareHM, Xu2020BuildingHH}, thereby boosting immersive experiences and improving the level of interactivity in human-computer interaction. Most of the existing 3D hand mesh recovery methods \cite{Choi2020Pose2MeshGC, Ge20193DHS, Boukhayma20193DHS} focus on coordinates associated with a predefined root position, \eg the wrist of a hand, and are unable to accurately determine the absolute camera-space coordinates of a mesh. This limitation hinders its applicability and makes it unsuitable for precise interaction tasks, \eg telemedicine surgery \cite{Lin2018AFM}. 

Camera-based 3D hand mesh recovery, which encompasses both 3D reconstruction and spatial localization, faces challenges due to the varied structures of the hand and the depth ambiguity present in RGB images. In most existing recovery methods ~\cite{Moon2020I2LMeshNetIP, Chen2021CameraSpaceHM, Iqbal2018HandPE, Hasson2019LearningJR, Hasson2021TowardsUJ}, a two-stage estimation scheme is typically employed. The first stage captures local structures by integrating key hand landmarks and their connections, while the second stage comprehends spatial semantic information. However, the networks used in these two stages in most existing methods operate independently and in a sequential mode, which can lead to unnecessary network overhead~\cite{Moon2020I2LMeshNetIP, Iqbal2018HandPE, Hasson2019LearningJR} and inefficiency during training ~\cite{Chen2021CameraSpaceHM}. Moreover, as most current methods either narrowly focus on local features or become overly fixated on the global context, they struggle to fully leverage the scale information inherent in the image. Therefore, the resistance to interference in a complex environment and the accuracy of coordinate positioning will be significantly reduced \cite{Moon2019CameraDT}.

This study proposes a camera-space 3D hand mesh recovery model in order to generate a precise and robust 3D hand mesh model in scenarios involving complex backgrounds and self-occlusion. By following a two-stage method, the hand mesh recovery is separated into root-relative and root-recovery tasks. The root-relative recovery task captures the relative positions of hand mesh vertices with respect to the hand's root node, while the root recovery task concentrates on determining the spatial position of the hand by using the bounding bin approach. Different from most existing methods, the networks used in the first and second stages are integrated as one in our model, which not only allows end-to-end training but also reduces the complexity. 

Our approach adopts an Encoder-Decoder architecture to obtain high-resolution 2D scale-aggregated features as foundational cues for 3D inference. This architecture has demonstrated effectiveness in various fields, including pose reconstruction and depth estimation \cite{Kim2022GlobalLocalPN, Xu2021GraphSH}. The well-established parameterized hand model, MANO \cite{Romero2017EmbodiedH} is utilized to enhance the inference. To perform nonlinear learning and leverage the relationships between vertices in the network topology for local hand shape inference and reconstruction, we employ spectral convolutional neural networks \cite{Defferrard2016ConvolutionalNN}. This approach effectively captures and utilizes the complex spatial relationships within the hand model. The depth estimation problem is formulated as a bin centers classification task to avoid the slow convergence and suboptimal local minima in training. Additionally, attention mechanisms \cite{Vaswani2017AttentionIA} are incorporated to guide the learning of hand features toward depth hierarchy information, even in the presence of complex backgrounds. By leveraging attention mechanisms, our model can selectively focus on relevant regions and effectively learn the necessary information. Finally, the collected information is combined to obtain the required 3D mesh information in camera space. This integration of information from various sources enables us to accurately reconstruct the 3D mesh representation of the hand. The performance of our proposed model is evaluated and compared with SOTA methods of root relative and camera space recovery methods by using one of the benchmark datasets, FreiHAND which includes numerous hand images with complex backgrounds and self-occlusion scenarios.

The main contributions of this work are as follows: 
\begin{itemize}
    \item Proposing a novel two-stage solution encompassing parallel processing for root-relative tasks and root recovery tasks, aiming to achieve camera-space hand pose estimation.
    \item Employing an implicit heatmap approach during the two-dimensional feature extraction stage for two-dimensional regression, to meet the requirements of the sub-tasks. This method enhances the effective collection of two-dimensional cues, creating conditions for end-to-end training of the network.
    \item Designing an Inception Graph Network based on the Inception architecture, which increases the receptive field of each convolution operation, enhancing the network's resistance to interference and robustness.
    \item Designing a method based on local detail and global feature attention for bin centers estimation, enabling robust root recovery task.
\end{itemize}

The rest of this paper is organized as follows. Section \ref{sec:relatedwork} discusses the concepts related to our model including hand recovery and depth estimation. Section \ref{sec:method} provides a detailed discussion of the content specifics within the three sub-modules encompassed by our model. Section \ref{sec:exp} outlines the experiments conducted on the FreiHAND dataset under standard evaluation conditions, along with implementation details. Section \ref{sec:conclusion} concludes this paper.

\section{Related Work}
\label{sec:relatedwork}

\subsection{Root-relative mesh/pose recovery}

The objective of root-relative mesh/pose recovery is to achieve accurate reconstruction of 3D mesh/pose according to a common root element. The methods can be mainly separated into three types: voxel-based~\cite{li2022detailed, Moon2020I2LMeshNetIP}, parameter-based~\cite{Zhou2020MonocularRH}, and coordinate-based methods. We focus primarily on coordinate-based methods.

Coordinate-based methods restore shapes by predicting the coordinates of each vertex. Recovering mesh/pose coordinates in such a high-dimensional space is highly challenging, with a key focus on the 2D-to-3D lifting~\cite{chen2023joint, Zhang2019PixelwiseR3}. This has led to the development of various spatial methods to achieve this goal. A spiral convolution approach handles mesh data in the spatial domain \cite{Lim2018ASA}. Graph-based methods can better leverage the relationships between vertices in the network topology for nonlinear transformations~\cite{kourbane2022graph}. Moreover, a hand mesh estimation network based on spectral filtering in graph convolutional networks is proposed \cite{Ge20193DHS}. Our model aims to explore the efficient mapping of graph neural networks for the 2D-to-3D transformation based on this approach.


\subsection{Root recovery}

Some mesh recovery methods with a single RGB image take a single-stage approach to directly obtain absolute coordinates. For example, intermediate vectors captures the spatial relationship between hand joints and densely sampled points, enabling implicit learning \cite{Huang2023NeuralVF}. On the other hand, a typical two-stage approach involves root-relative pose prediction and root depth reconstruction. Previous studies have treated these tasks as separate components, employing independent networks with limited interdependencies. For instance, after predicting the 2.5D pose, 2D dense regression estimates the 3D hand pose \cite{Iqbal2018HandPE}. However, reconstructing the absolute three-dimensional hand pose in camera space requires an additional estimation of the global scale. 1D dense regression recovers the root-relative 2.5D pose, while RootNet \cite{Moon2019CameraDT} is used to predict the root's absolute coordinates in camera space, enabling the estimation of the hand pose \cite{Moon2020I2LMeshNetIP}. A method combining the generated root-relative mesh and 2D information has been proposed for root coordinate registration using quadratic programming \cite{Chen2021CameraSpaceHM}.


\subsection{Bin centers method in depth recovery field}
The task of depth reconstruction from RGB images is inherently challenging due to its ill-posed nature. Issues such as limited scene coverage, scale ambiguity, and the presence of translucent or reflective materials can result in ambiguous geometric shapes that cannot be directly inferred from appearance alone. Regression-based methods often suffer from slow convergence and local suboptimal solutions. To address these challenges, an alternative approach is proposed, where the depth estimation problem is transformed into a classification task. Fu et al. suggest dividing the depth range into a fixed number of pre-defined bins with predetermined widths and formulate depth perception as the prediction of confidence scores for different positions within these bins \cite{Fu2018DeepOR}. This approach is combined with graph neural networks to enable effective feature propagation between joints, specifically addressing the root joint localization problem in multi-person 3D pose estimation in camera space \cite{Lin2020HDNetHD}. The computation of adaptive bins dynamically adjusts based on the input scene characteristics. They leverage the advantages of depth map regression by predicting the final depth value as a linear combination of bin centers \cite{Bhat2020AdaBinsDE}. In this work, our objective is to investigate the effective utilization of the bin centers method, considering both local details and global features, for the purpose of hand root recovery.


\section{Methodology}
\label{sec:method}

\begin{figure*}[t]
\centering
\includegraphics[scale=0.7]{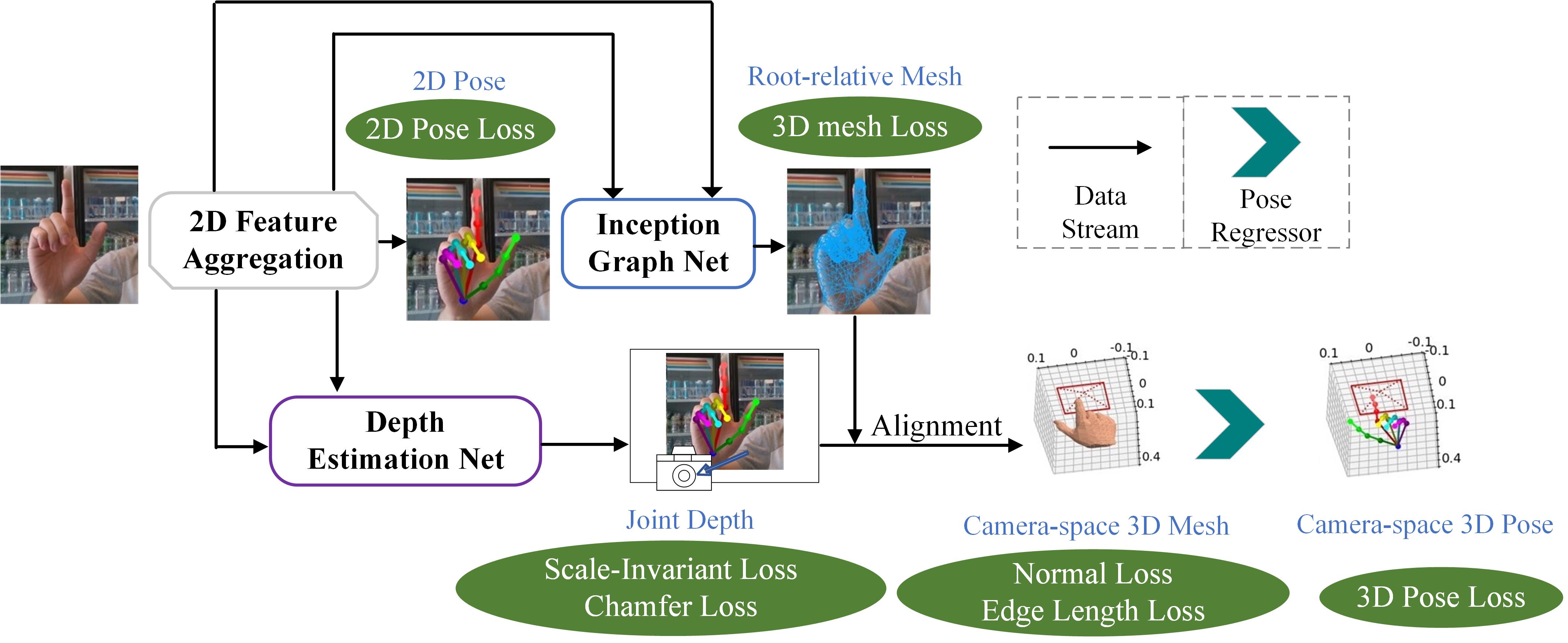}
\caption{Structure of our proposed model.}
\label{fig:flow}
\end{figure*}

Our two-stage model aims to recover a 3D hand mesh in the camera space from a single image. The procedures of our model are illustrated in Figure \ref{fig:flow}. The 2D global aggregation features and 2D joint alignment features are first extracted from a given RGB image by a deep convolutional neural network (CNN) in the process of 2D feature aggregation (Section \ref{sec:method:feat}). The information is used to support the process of the stages of the root-relative 3D mesh prediction and root depth reduction. To improve the efficiency and accuracy of 2D information, the approach incorporates implicit learning of heatmaps and utilizes the method of 2D coordinate relay supervision. The combination allows the extracted information to be more beneficial for the subsequent stages of the process. The 3D mesh prediction method captures the variations in hand structure. Our model revises the Res-Inception spectral graph convolution module. By incorporating neighboring information from different receptive field sizes, the model's capability is improved to integrate spatial information across nodes. This enhancement contributes to a more comprehensive understanding of the spatial relationships within the hand structure. The objective of root depth estimation is to precisely estimate the distance between the root of a hand and each vertex. The objective of root depth estimation is to precisely estimate the 3D position of the hand's root in the camera space. The images are dynamically divided into depth intervals, and a classification approach is used to determine the final depth values. In order to fully leverage the information within the images, attention-based interaction is employed to integrate global background information with local hand-specific details. This reduces the depth ambiguity in the model and enhances the accuracy of root depth estimation.

\subsection{2D feature aggregation}
\label{sec:method:feat}

The successful gathering of 2D information is key to making accurate forecasts in the following stages of our model. A CNN is employed for feature extraction due to its strong generalization capability in 3D estimation tasks \cite{He2015DeepRL}. The objective of the 2D feature aggregation is to parse the 2D information in a way that fulfills both the root-relative prediction and absolute depth prediction stages, which also impose new requirements on the model's intermediate representation or additional supervision. 

\begin{figure*}[t]
\centering
\includegraphics[scale=0.7]{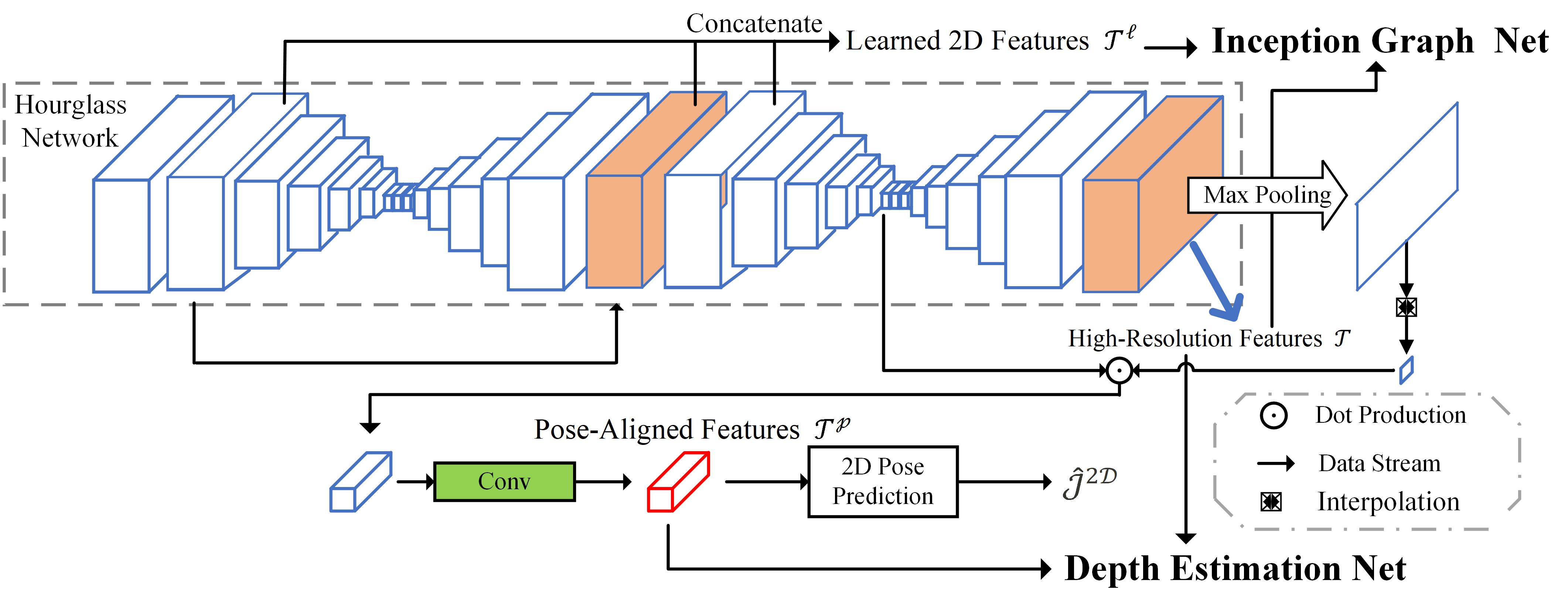}
\caption{Schematic diagram of the 2D feature aggregation.}
\label{fig:2d_graph}
\end{figure*}

The procedure is illustrated in Figure \ref{fig:2d_graph}. Inspired by~\cite{Chen2021MobReconMH, Nibali2018NumericalCR}, a hybrid method combining implicit learning of heatmaps with regression of 2D coordinates is proposed in our model. Although heatmaps are estimated in most existing feature aggregation methods, \eg the Hourglass network~\cite{Newell2016StackedHN}, gradient flows generated around the nodes lead to the loss of edge details in the encoded features and hinder the penetration of local and global information understanding. Another kind of feature aggregation method, named direct regression \cite{Toshev2013DeepPoseHP}, predicts joint information of the 2D hand pose through global semantics and receptive fields. However, direct regression may suffer from a loss of local semantics since it allows for significant preservation of the complete spatial semantic information learned by the feature maps themselves through regression constraints. Consequently, our model integrates both types of methods to enhance the quality of extracted features.

Our model contains a multi-layer Hourglass network to generate high-resolution feature maps $\mathcal{T} \in \mathbb{R}^{B \times C_t \times h_t \times w_t}$, where $B$ denotes the batch size of learning, $C_t$ represents the length of the feature vectors, $h_t$ and $w_t$ denote the width and length of $C_t$ respectively. In order to alleviate memory pressure, $h_t = H / 2$ and $w_t = W / 2$ are selected, enabling enhanced learning with larger batch sizes, where $H$ and $W$ are representative of the height and width of the original image respectively. Instead of adding heatmap constraints to the feature map $\mathcal{T}$, we input it into a max pooling layer for pose pooling~\cite{Zhang2021InteractingT3}. The feature map undergoes downsampling through bi-linear sampling and is denoted as $\mathcal{T}^d$. The pose-aligned feature $\mathcal{T}^p$ is extracted from the lowest level of the skip connections as follows:
\begin{equation}
    \mathcal{T}^p = \Theta(\mathcal{T}^d\odot interpolation(\mathcal{T}))
\label{eq:featmap}
\end{equation}
where $\Theta$ represents a convolutional layer, $\odot$ denotes a dot production, and interpolation is a process of bilinear downsampling. A $1 \times 1$ convolutional layer aligns the feature channel dimensions with the number of joints. The dimensions of $\mathcal{T}^p$ are $h_e \times w_e$. By focusing on the 2D space of pose-aligned features instead of global features, $\mathcal{T}^p$ is able to maintain the local details of the image. 

The last two dimensions of the pose-aligned features are transformed $\mathcal{T}^p$ into a flattened representation with dimensions $\mathbb{R}^{B \times C_t \times (h_e \times w_e)}$. $\mathcal{T}^p$ are fed into the stage of depth estimation, and also the 2D pose prediction. Our model uses a Multi-Layer Perceptron (MLP) Neural Network to estimate the 21 key points $\hat{J}^{2D}$ in 2D pose prediction as intermediate supervision. The MLP model consists of two fully connected layers and an activation layer, with a specific sequence: a $(h_e \times w_e) \times 32$ fully connected layer, an activation layer, and a $32 \times 2$ fully connected layer.

\subsection{Root-relative 3D mesh prediction}
\label{sec:method:mesh}

\begin{figure}[t]
\centering
\includegraphics[width=3.4in]{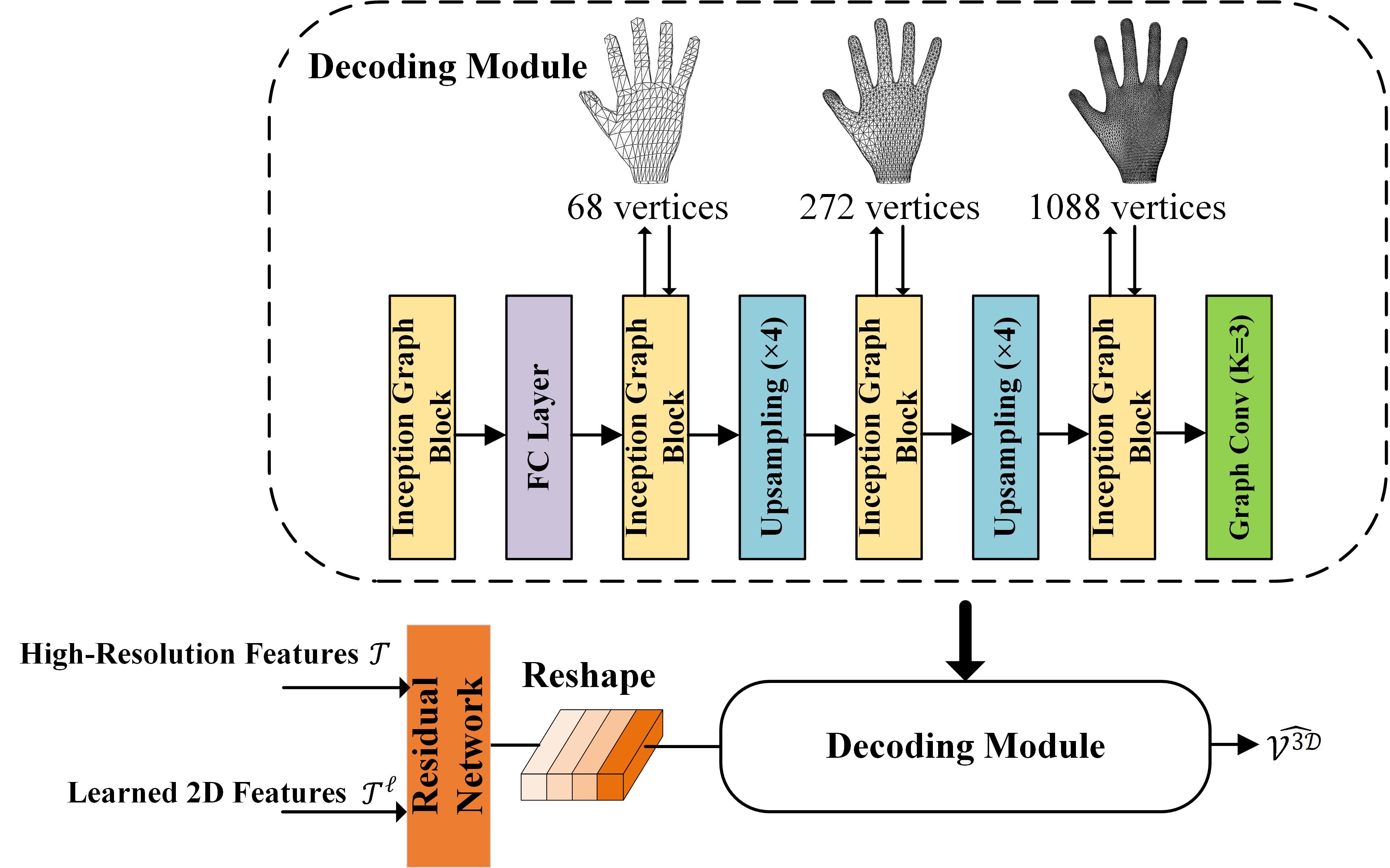}
\caption{Structure of Inception Graph Network.}
\label{fig:meshprediction}
\end{figure}

\begin{figure}[t]
\centering
\includegraphics[width=3.4in]{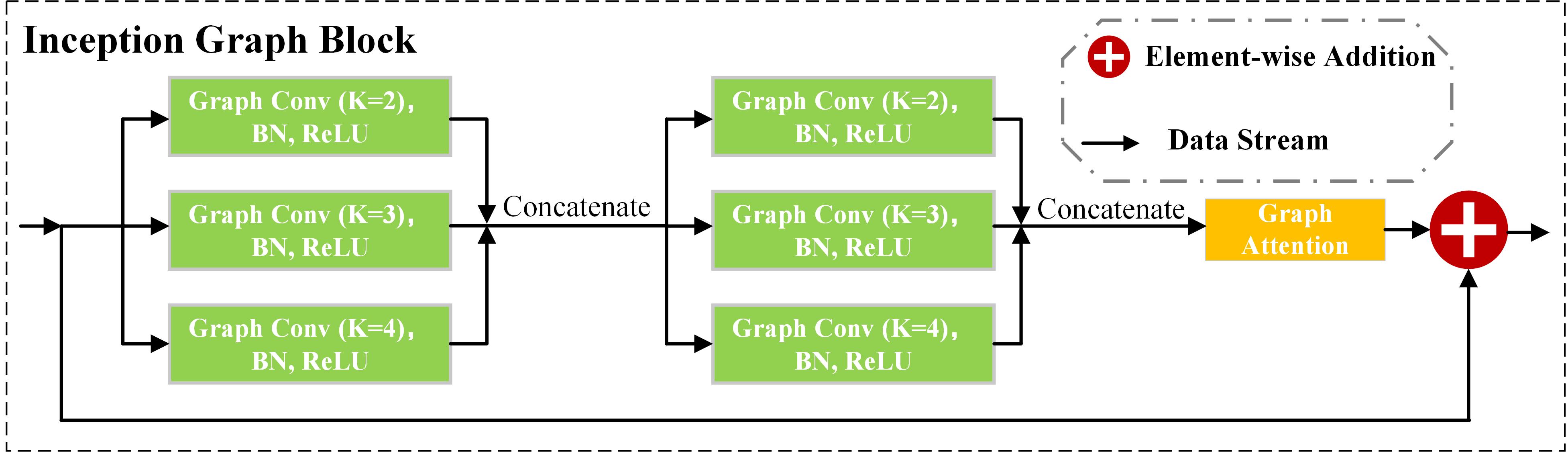}
\caption{Structure of our proposed Inception Graph Block.}
\label{fig:incepblk}
\end{figure}

The reconstruction of hand shapes from RGB images is a critical task in 3d hand mesh recovery. The objective of this stage is to effectively address the spatial gap between 2D poses and 3D forms by accurately inferring the plausible 3D hand structure, even in complex environments or under self-occluded conditions. The spatial gap refers to the disparity between the pixel-level representation of hand poses provided by 2D images, and the complete information regarding the true form and position of hands in the 3D space that they do not provide. To address this, we propose the utilization of a spectral graph convolutional network, specifically the Inception Graph Net. By adopting a coarse-to-fine approach and predicting a dense mesh of the hand, we can achieve a relative prediction of the hand shape. In the following sections, we will introduce the different components of our proposed Inception Graph Net in a sequential manner.

The residual network first converts the high-resolution feature $\mathcal{T}$ along with the learned 2D features $\mathcal{T}^l$ into latent feature vectors $\mathcal{T}^a$ to further enhance the representation capabilities of the features.
The latent vectors are then reshaped and processed by the decoding module. The objective of the module is to refine an initial 3D mesh model based on the 2D information obtained in the previous step, in order to achieve a more precise hand shape structure. A 3D hand mesh \cite{Romero2017EmbodiedH} is represented by an undirected graph $\mathcal{M} = (\mathcal{V}, \mathcal{E})$ characterized by its vertices $\mathcal{V} = \{v_i\}_{i=1}^{N}$ and edges $\mathcal{E} = \{e_i\}_{i=1}^{E}$, and contains 778 vertices and 3187 edges. A triangular mesh model is applied to present a hand shape.

Our decoding module shown in Figure \ref{fig:meshprediction} mainly follows the model proposed in \cite{Ge20193DHS}. To perform graph coarsening on the predefined hand graph $\mathcal{M}$, a topological calculation is carried out. This process generates three different resolutions of balanced binary tree structure graphs, following the approach presented in the weighted graph coarsening method \cite{Dhillon2007WeightedGC}. These fixed graphs are fed into the training process. Gradual upsampling obtains a set of vertices $\hat{\mathcal{V}}^{3D}$ for the hand mesh. 

Similar to \cite{Ge20193DHS}, our model includes a fully connected layer, two upsampling layers, and a standalone graph convolutional layer. However, Our proposed Inception Graph Block is used to extract the features, rather than relying on a graph convolutional layer. The graph convolutional layer is K-localized. As the convolution corresponds to a Kth-order polynomial in the Laplacian matrix, only the nodes within a K-step distance from the center node (K-hop neighborhood) are considered ~\cite{Defferrard2016ConvolutionalNN}. A single $K$-hop is used in current graph convolutional networks ~\cite{Ge20193DHS, Choi2020Pose2MeshGC}, treating fixed-range neighboring vertices indiscriminately. A fixed receptive field of the convolutional operation is typically sufficient for standard hand pose estimation tasks. However, the presence of interfering information presents a notable challenge for the network, limiting its ability to accurately infer ambiguous information within a limited context.

In order to enhance the adaptability of the network in dealing with complex environments and self-occlusion scenarios, an Inception Graph Block is proposed to replace the graph convolutional layer. The block considers multiple $K$-hop neighborhoods. By enabling the learning of correlations among nodes within various neighborhood ranges, the model becomes capable of making rational inferences for ambiguous information. This is achieved by leveraging the positional information of related nodes at the same neighborhood level. As shown in Figure \ref{fig:incepblk}, three stacked Graph Convolutional Networks with different K-hop convolutions are employed twice. By aggregating information from various local neighbors, the spectral graph convolution enhances its learning ability and expands the receptive field. This enables the model to capture and incorporate more comprehensive information during the convolutional operation. It also improves the perception of correlations among different receptive fields, thereby enhancing the network's reasoning abilities in diverse environmental contexts. This exploration of feature interactions between different receptive fields enables the model to collectively adapt to various local features.

SENet~\cite{Hu2017SqueezeandExcitationN} used as a Graph Attention layer is concatenated at the end of the Inception Graph block. This allows the model to focus on learning after information aggregation, enhancing the perception capability between edge nodes and emphasizing feature channels with more significant node information. As a result, the model can better understand and learn multiscale information effectively.


\subsection{Depth Estimation}

\begin{figure}[!t]
\centering
\includegraphics[scale=0.48]{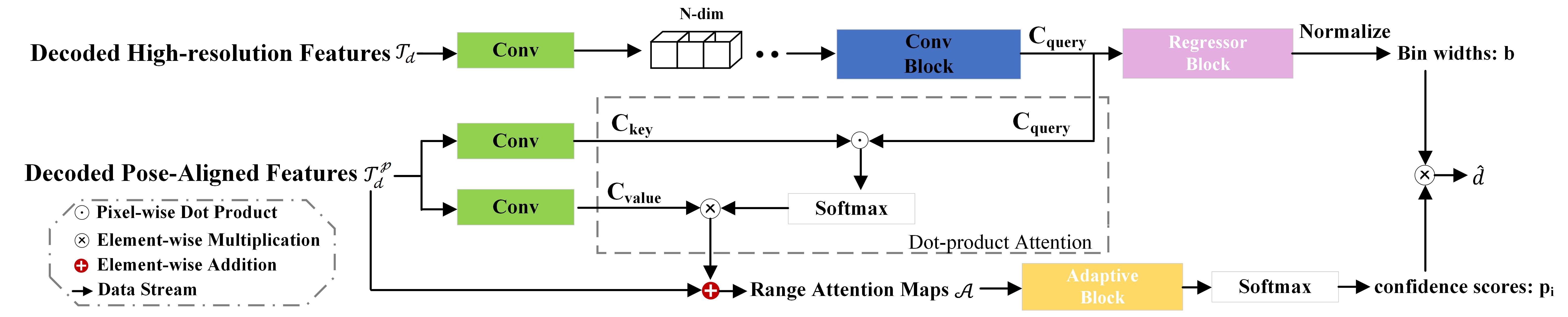}
\caption{Structure of Depth Estimation Net.}
\label{fig_depth}
\end{figure}

The depth estimation process outputs the distance from the root for each vertex of the mesh model. Similar to existing studies ~\cite{Fu2018DeepOR}, our model formulates this regression problem as a classification task. The actual depth of a hand tends to be within a narrow range due to its size. The limited range is partitioned into various bin intervals. Vertices are categorized into these bins according to their depth information. Due to the lack of depth reference information in hand images, an attention mechanism is embedded in the depth estimation of our model. 

The depth range $(\hat{d}_{min}, \hat{d}_{max})$ is first estimated according to ~\cite{Lin2020HDNetHD}: 
$\hat{d} = \frac{d}{\sqrt{f_x \times f_y}}$,
where $f_x$ and $f_y$ represent the camera focal lengths in the $x$ and $y$ directions respectively, and $d$ represents the actual depth value. Our objective is to achieve efficient depth segmentation while minimizing the impact of camera parameters.

The specific network structure is depicted in Figure~\ref{fig_depth}. Specifically, decoding operations are performed on the high-resolution feature $\mathcal{T}$ and pose-aligned features $\mathcal{T}^{p}$ separately through two $1 \times 1$ convolutional layers. This resulted in decoded high-resolution feature $\mathcal{T}_{d}$ and decoded pose-aligned features $\mathcal{T}_{d}^{p}$. The decoded high-resolution feature $\mathcal{T}_{d}$ is embedded by using a convolution layer with a $s \times s$ convolution kernel and $s$ strides to produce a tensor of size $h_t/s \times w_t/s \times S_E$, where $S_E$ is the embedding size, $h_t$ and $w_t$ denote the size of $\mathcal{T}_{d}$. The tensor is fed into the Conv Block to explore the spatial semantics of the image, and the output is reshaped into individual flattened tensors $x_p \in \mathbb{R}^{L \times 1 \times S_E}$, where $L=\frac{h_tw_t}{s^2}$. The sequence consisting of the first N-Query tensors is referred to as the spatial query factor $C_{query}$. It should be noted that $C_{query}$ contains global hierarchical information of depth for each image and also associated semantic information across channels. This causes $C_{query}$ different from its patch embedding counterpart~\cite{Bhat2020AdaBinsDE}.

The depth range needs to be divided into $N$ bins, where $N$ is a pre-defined positive integer. A series of one-dimensional convolution form a Regressor Block that operates on $C_{query}$. Its purpose is to derive a vector $y$ of dimensionality $N$. The width of the $i$th bin denoted by $b_i$ is calculated by normalizing $y$ as follows:
\begin{equation}
    b_i = \frac{y'_i + \epsilon}{\sum_{i=1}^{N}y'_j + \epsilon},
    \label{eq:normalize_b}
\end{equation}
where $\epsilon$ represents a small positive number aiming to ensure a positive bin width. The center of a bin is determined based on vector $b$ obtained by normalizing the $y$ vector by:
\begin{equation}
    B(b_i) = \hat{d}_{min} + (\hat{d}_{max} - \hat{d}_{min})(b_i / 2 + \sum^{i - 1}_{j=1}b_j),
    \label{eq:bin_centers}
\end{equation}

The existing depth estimation methods primarily concentrate on the global depth reconstruction of images, and may not perform optimally when it comes to specific objects within the images, thereby compromising the accuracy of object-level precision. The arrangement of joints holds valuable data for expression. This encompasses details about the spatial layout of the internal structure, such as the distances between joints and the angles they form. Additionally, there are complex biomechanical coordination relationships among hand joints that offer essential clues for determining the main joint. Hence, the inclusion of joint information facilitates the seamless integration of two-dimensional joint extraction, ultimately reducing potential ambiguity. Our proposed attention residual learning mechanism uses soft attention masks to guide the network to focus on the relationship between the hand joint local information and the hand's global spatial information. The confidence $p_i$ of each the $i$th bin $b_i$ is estimated.

The extraction of deep hierarchical information from global image features plays a crucial role in making local object judgments. In order to accomplish this, the pose-aligned local features $\mathcal{T}^p$ serve as the key value for cross-attention, while the spatial query factor $C_{query}$ acts as the query. By employing attention mechanisms, the global and combined local relational information is then aggregated into the pose mapping features. This enhances the network's perceptual capabilities, allowing for more accurate and comprehensive analysis. To be specific, the decoded 2D joint-aligned features $\mathcal{T}^{p}_{d}\in \mathbb{R}^{B \times E \times W \times H}$ are transformed by two $1\times1$ convolutional blocks with a stride of 1 to the key-value feature pairs $C_{key}$, $C_{value} \in \mathbb{R}^{B \times E \times W \times H}$. Each input segment is associated with a specific $C_{key}$, which is used to determine the relevance between $C_{query}$ and various segments of global information. On the other hand, $C_{value}$ is employed to compute the weighted sum of outputs, with the weights being determined by the level of association between $C_{value}$
 and $C_{key}$.

$C_{key}$, $C_{value}$ and $C_{query}$ are fed into the dot-product attention module. The utilization of attention mechanisms enables our model to concentrate on essential details during the process of reconstructing depth. Consequently, this enhances both the performance and understanding capabilities of the model. However, the transformation of the attention mask from 0 to 1 may result in a reduction of feature values, which makes the model vulnerable to the noise introduced by the main features. Using the attention modules alone may lead to performance degradation. To mitigate this issue, the residual learning is considered in the attention module. Residual learning involves each layer learning the residual in relation to the preceding layer, rather than solely focusing on learning the mapping of the input. This approach helps address the issue of significant degradation in feature values. By incorporating residual learning, the problem of excessive degradation can be alleviated. The range attention maps $\mathcal{A}$ is obtained. $\mathcal{A}$ is then aligned with the channels of the root node index and $N$ depth bin centers through convolutional layers and adaptive average pooling layers. A softmax activation is applied to obtain the confidence scores $p_i$ for the $i$th bin center, where $i = 1..N$. The final depth value $\hat{d}$ is computed as a linear combination of the confidence scores $p_i$ and the corresponding N depth bin centers $B(b)$, as follows:
\begin{equation}
    \hat{d} = \sum\limits^{N}_{i=1} B(b_i)p_i,
    \label{eq:depth_compute}
\end{equation}

The final depth values are obtained by performing inverse computations with the camera parameters. 


\section{Loss Fuctions}
\label{sec:loss}

Our model is optimized by minimizing \eqref{loss:overall_loss} in an end-to-end manner. The overall loss function is defined as follows:
\begin{equation}
    \begin{aligned}
        \mathcal{L}_{total} = \lambda_{p2d}\mathcal{L}_{p2d} + 
        \lambda_{d}\mathcal{L}_{d} + \lambda_{b}\mathcal{L}_{b} + \lambda_{m}\mathcal{L}_{m} \\
        + \lambda_{p3d}\mathcal{L}_{p3d} + \lambda_{n}\mathcal{L}_{n} + \lambda_{e}\mathcal{L}_{e},
    \end{aligned}
    \label{loss:overall_loss}
\end{equation}
where $\mathcal{L}$ denotes the loss while $\lambda$ represents its associated weight. $\lambda$ serves as a hyperparameter that balances the influence of the respective loss terms in the overall loss function. 

The 2D feature extraction stage uses 2D poses as intermediate supervisory targets for the first phase. The aligned 2D features obtained are input into a convolutional module and fully connected layers to generate predicted 2D poses. The loss ${L}_{p2d}$ is represented by the L1 norm of the Euclidean distance between the model's predicted 2D hand pose and the corresponding ground truth on the pixel plane.
\begin{equation}
    \mathcal{L}_{p2d} = ||\mathcal{\hat{J}}^{2D} - \mathcal{J}^{2D(gt)}||_1, 
    \label{loss:j2d}
\end{equation}
where $\mathcal{J}^{2D(gt)} \in \mathbb{R}^{21 \times 2}$ represents the ground truth values for each 2D hand keypoint in $\mathcal{J}^{2D}$, and $\hat{}$ represents the predicted values.

The scaled version~\cite{Lee2019FromBT} of the Scale-Invariant (SI) loss is applied to estimate the depth value for hand key points:
\begin{equation}
    \mathcal{L}_d = \sqrt{\frac{1}{N}\sum\limits_{i}g_{i}^2 - \frac{\lambda}{N^2}(\sum\limits_{i}g_{i})^2},
    \label{loss:SI}
\end{equation}
where $g_i = log\hat{d_i} - log{d_i}$ and $d_i$ is the ground truth depth for $i$ hand key point. $N$ is the number of hand key points with valid ground truth depth values.

Since the distribution of bin centers predicted by the network should follow the distribution of depth values in the ground truth, and vice versa, the bi-directional Chamfer Loss ~\cite{Fan2016APS} is used as a regularizer:
\begin{equation}
    \mathcal{L}_b = chamfer(X, c(b)) + chamfer(c(b), X), 
    \label{loss:chamfer}
\end{equation}
where $c(b)$ denotes the set of bin centers and $X$ represents the set of all depth values in the ground truth image.

One possible regression coefficients for the root-relative pose joints are provided by MANO \cite{Romero2017EmbodiedH}. The loss term based on the Euclidean distance between them and their corresponding ground truth is calculated using the L1 norm:
\begin{equation}
    \mathcal{L}_v = ||\hat{\mathcal{V}}^{3D} - \mathcal{V}^{3D(gt)}||_1, \mathcal{L}_{p3d} = ||J\hat{\mathcal{V}}^{3D} - \mathcal{J}^{3D(gt)}||_1,
    \label{loss:j3d}
\end{equation}
where $\mathcal{V}^{3D(gt)} \in \mathbb{R}^{778 \times 3}$ represents the ground truth values for each root-relative hand mesh vertex in $\mathcal{V}^{3D}$, and $\mathcal{J}^{3D(gt)} \in \mathbb{R}^{21 \times 3}$ represents the ground truth values for each root-relative 3D hand key point in $\mathcal{J}^{3D}$. The variable $J$ refers to the pose regression coefficient.

Simultaneously, to enhance the fidelity of fitting the mesh to reality and mitigate the risk of getting stuck in local minima. The normal loss $\mathcal{L}_n$ is incorporated to reinforce surface normal consistency:

\begin{equation}
    \mathcal{L}_{n} = \sum\limits_{f\in\mathcal{F}}\sum\limits_{(i, j)\subset f}|\frac{\hat{\mathcal{V}}_i^{3D} - \hat{\mathcal{V}}_j^{3D}}{||\hat{\mathcal{V}}_i^{3D} - \hat{\mathcal{V}}_j^{3D}||_2}\cdot n_f^{(gt)}|,
    \label{loss:normal}
\end{equation}
where $f$ is an index representing a triangle face $\mathcal{F}$ within the hand mesh and $(i,j)$ represents the indices of the vertices forming an edge within the triangle $f$. $n_f^{gt}$ is the ground truth normal vector of the triangle face $f$, calculated from the ground truth vertices $V^{3D(gt)}$.

The edge length loss $\mathcal{L}_e$ is introduced to enhance edge length consistency:
\begin{equation}
    \mathcal{L}_{e} = \sum\limits_{f\in\mathcal{F}}\sum\limits_{(i, j)\subset f}|||\hat{\mathcal{V}}_i^{3D} - \hat{\mathcal{V}}_j^{3D}||_2^2 - ||\mathcal{V}_i^{3D(gt)} - \mathcal{V}_j^{3D(gt)}||_2^2|,
    \label{loss:edge_length}
\end{equation}


\section{Experiments}
\label{sec:exp}

\subsection{Experimental Settings}

Our proposed method is evaluated by using the FreiHAND dataset ~\cite{Zimmermann2019FreiHANDAD}, which comprises real hand color images captured using multi-view cameras. It contains 130,240 training samples and 3,960 evaluation samples collected from 32 individuals of various genders and racial backgrounds. The predicted results are submitted to the official server platform for online evaluation. During the training phase, the images are resized to a resolution of $224 \times 224$. Moreover, the data augmentation technique is applied to enhance generalization, including image scaling, rotation, translation, and color jitter. 

To further validate the generalization performance of our model, we attempted to utilize an alternative dataset - the Rendered Hand Pose Dataset (RHD) \cite{Zimmermann2017LearningTE}. It comprises 41,258 and 2,728 synthetic rendered samples for training and testing hand pose estimation, respectively. We employed the result trained on the FreiHAND dataset for fine-tuning testing on the RHD training set. Given that the RHD dataset does not provide hand mesh information in camera space, it serves solely as a measure of our model's relative pose prediction performance.

We found relatively fewer efforts in hand mesh recovery in camera space, so the latest year for statistical results is 2021. It is important to emphasize that the focus of the article is on an end-to-end method for the restoration of global coordinate information, which facilitates algorithm deployment and brings more convenience. End-to-end training cannot guarantee outstanding performance at each stage; therefore, the relative predictive results are only partially referential. Our hand mesh prediction results perform well in camera space comparisons on the FreiHAND dataset, which is significant.

ResNet \cite{He2015DeepRL} is used as the backbone of our model and its model parameters are initialized according to the Kaiming-normal distribution. The Adam optimizer \cite{Kingma2014AdamAM} with a batch size of 32 is used for network training. The fixed step decay with a total of 50 iterations is used as our training strategy. The initial learning rate is set at $10^{-4}$ and is divided by 10 at the 40th epoch. A personal computer with a single NVIDIA RTX 3090 is used in our experiments. All programs are implemented in PyTorch.

The parameters of the loss functions in our model initially follow \cite{Chen2021CameraSpaceHM, Bhat2020AdaBinsDE} and are adjusted during the training to maximize the performance. The following settings are recommended: $\lambda_{p2d} = 1, \lambda_{d} = 10, \lambda_{b} = 10, \lambda_{m} = 1, \lambda_{p3d} = 1,  \lambda_{n} = 0.1, \lambda_{e} = 1$.

The average errors on the estimated root-relative joint (CV) and vertex (PV) positions, and errors on the estimated joint (CJ) and vertex (CV) positions in the camera space are measured by the Euclidean distance. Moreover, the area under the percentage of correct key points (PCK) with various error thresholds is also considered. The average of performance on five-time independently is calculated in order to provide a reliable and robust evaluation.


\subsection{Comparison with state-of-the-art methods}

\begin{figure}[!t]
\centering
\subfloat[\label{pck_cv}]{\includegraphics[scale=0.27]{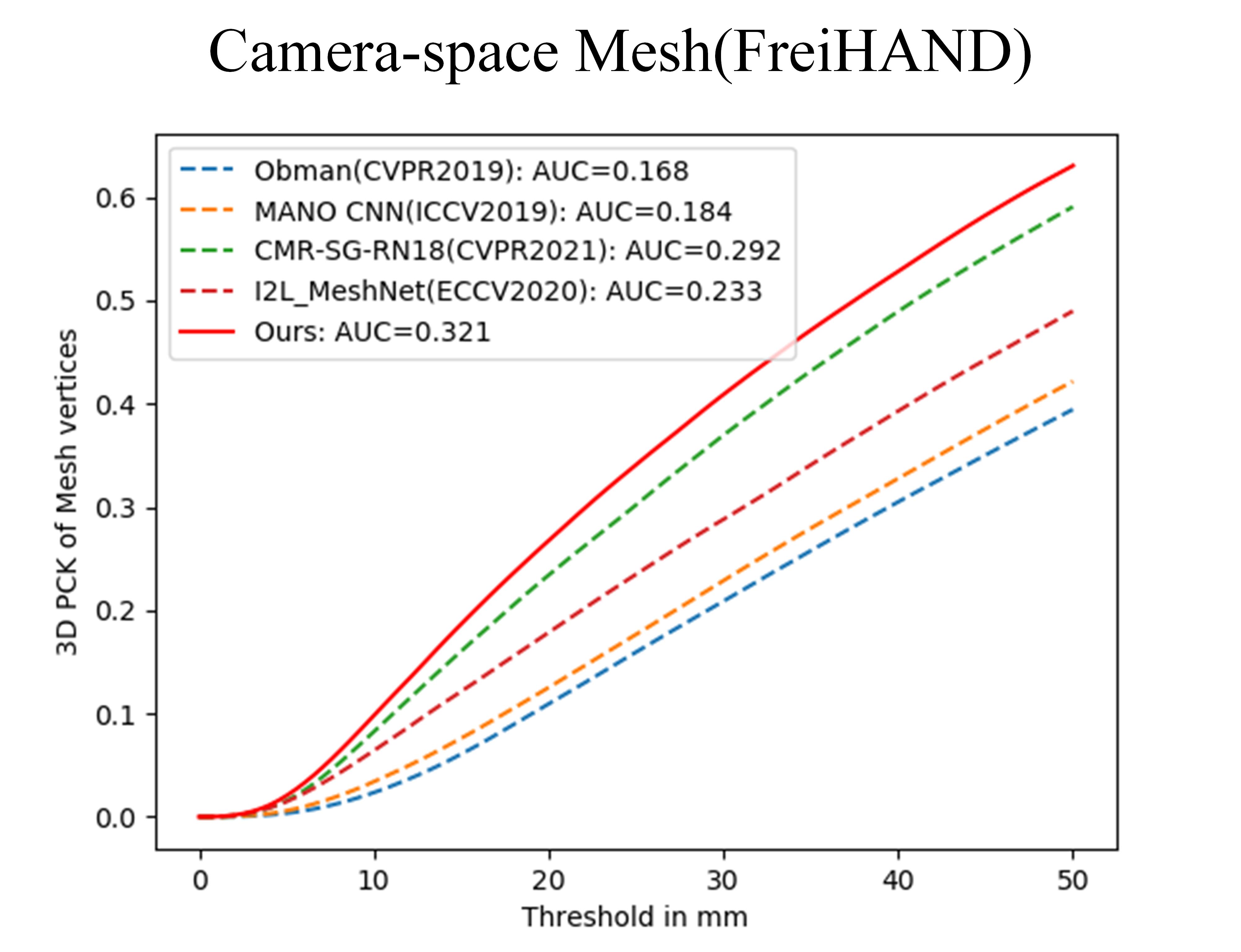}}
\subfloat[\label{pck_pj}]{\includegraphics[scale=0.27]{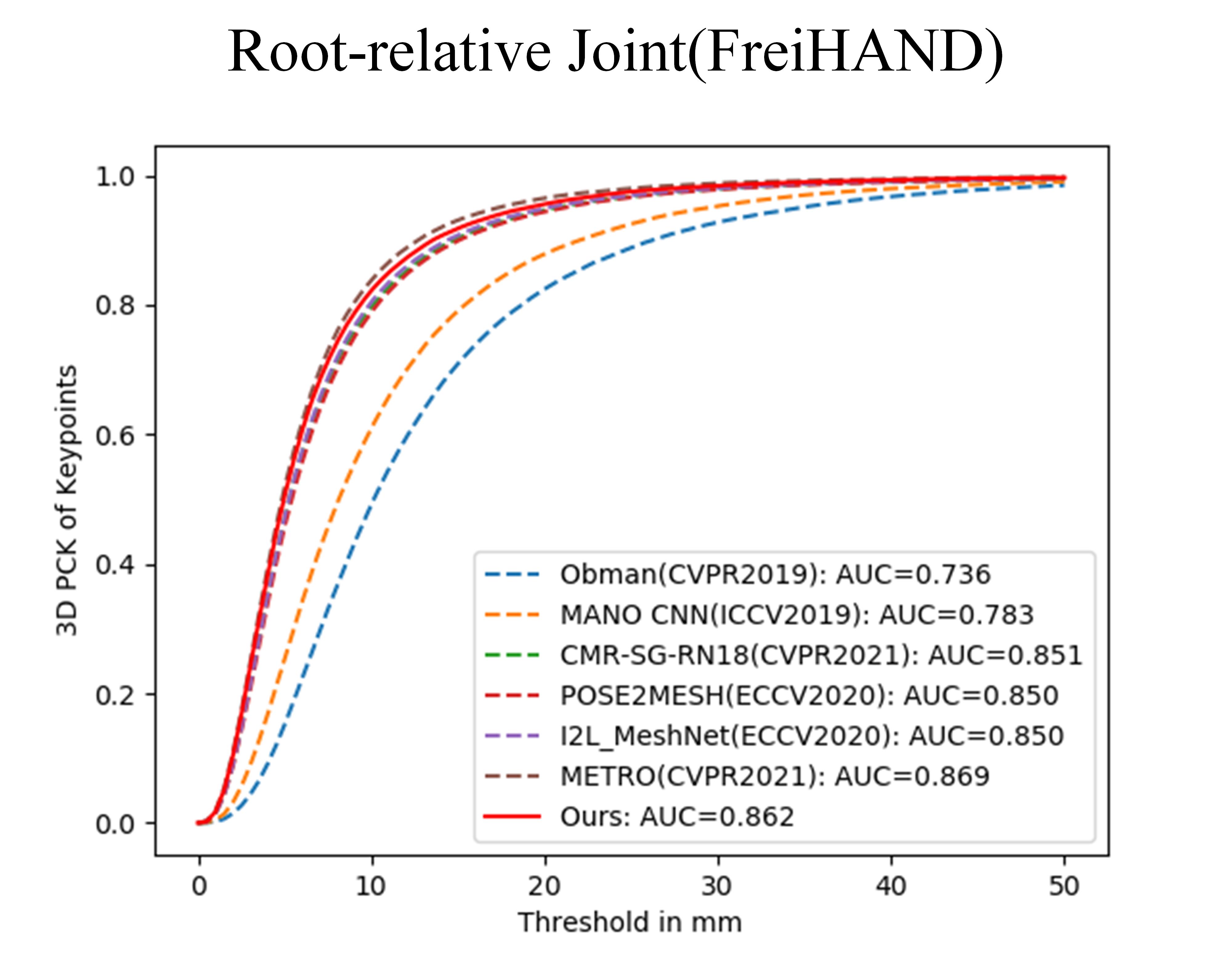}}
\subfloat[\label{pck_rhd}]{\includegraphics[scale=0.24]{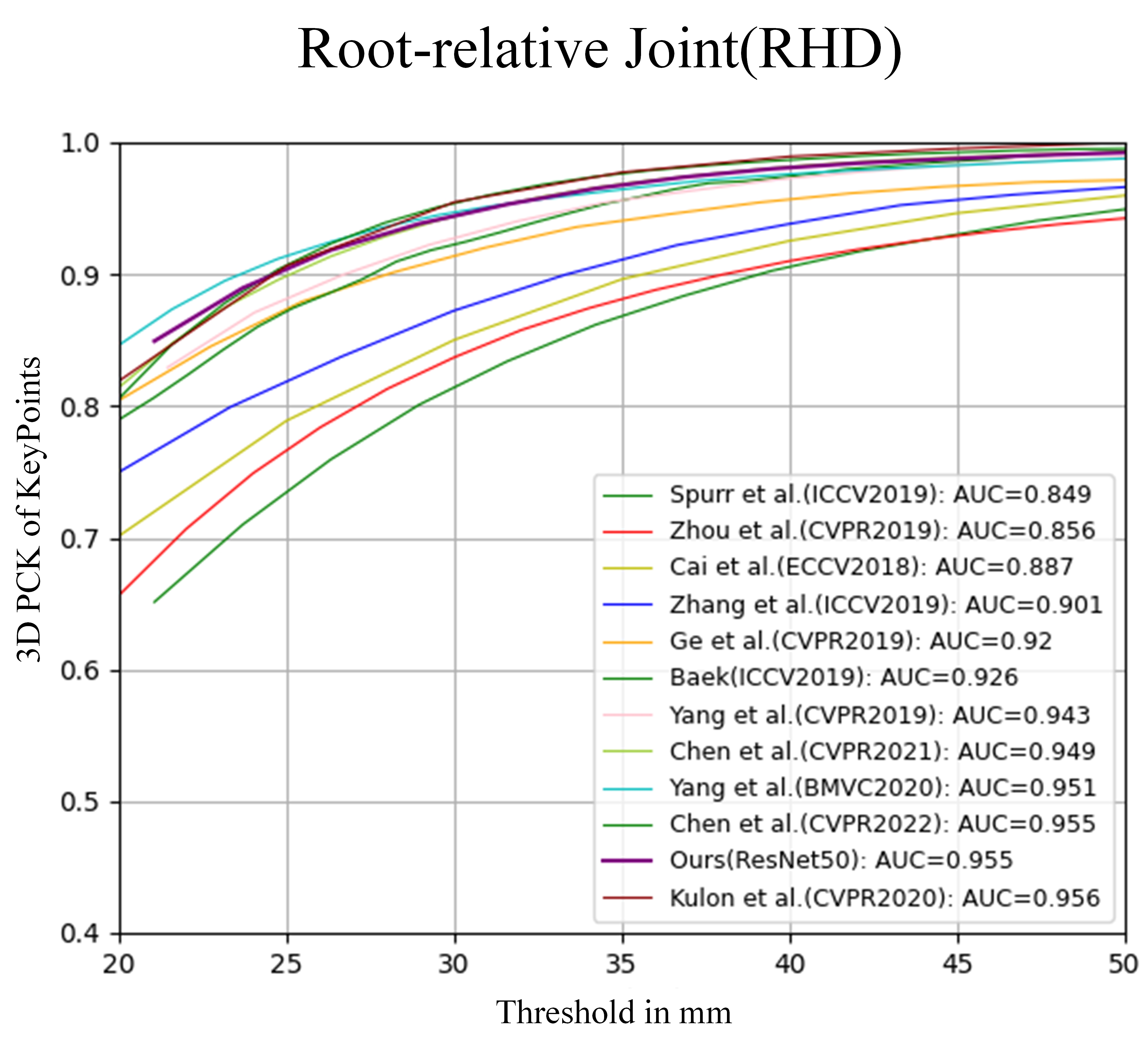}}
\caption{PCK vs. error thresholds.}
\label{fig:pck}
\end{figure}

The performance of our model is compared with state-of-the-art (SOTA) methods of 3D hand mesh recovery on FreiHAND. To ensure a fair comparison, all methods were configured under the same evaluation conditions and standards. The results of our proposed model compared with other SOTA methods are presented in Table \ref{tab:table1}. The results of all methods, except for our proposed method, are obtained by directly copying them from their respective studies.

When evaluating the results of root-relative recovery, METRO \cite{Lin2020EndtoEndHP} stands out as the top performer among all methods, achieving PJ and PV scores of 6.7 and 6.8 respectively. The excellent performance of METRO may be because its complex models with a large number of parameters. CMR-PG also delivers satisfactory performance in terms of PJ and PV, slightly trailing behind METRO. Spiral convolution \cite{Lim2018ASA} of CMR-PG has been shown to have an advantage over spectral graph convolution methods when handling locally structured data with specific patterns. However, for our study, it is crucial to consider global-local correlations and a wider range of contextual information. Therefore, spectral graph convolution is deemed more applicable and suitable for our work. Our method secures the third position and exhibits similar performance to CMR-PG. The root relative recovery methods, except Boukhayma one, perform better than the camera space methods in general.

In terms of estimation performance in the camera space, our method demonstrates exceptional results, surpassing all other methods evaluated. Our model achieves a CJ score of 47.1 and a CV score of 47.1. Notably, our method outperforms the second-best approach, CMR-PG, by a significant margin of 1.2 and 1.3 in CJ and CV respectively.

\begin{table*}[!t]
\label{tab:table1}
\caption{Recovery performance on FreiHAND.`PJ', `PV', `CJ', `CV' denote PAMPJPE, PA-MPVPE, CS-MPJPE, and CS-MPVPE, respectively, while `RR' and `CS' denote Root Relative and Camera-Space Methods.}
\centering
\resizebox{0.65\textwidth}{!}{\begin{tabular}{ c | c | c c c c}
\hline
Method & Type & PJ &PV & CJ & CV \\
\hline
Boukhayma \cite{Boukhayma20193DHS}& RR & 35.0 & 13.2 & - & - \\
Pose2Mesh \cite{Choi2020Pose2MeshGC} & RR & 7.7 & 7.8 & - & - \\ 
METRO \cite{Lin2020EndtoEndHP} & RR & \textbf{6.7} & \textbf{6.8} & - & - \\
\hline
ObMan \cite{Hasson2019LearningJR} & CS & 13.3 & 13.3 & 85.2 & 85.4 \\ 
MANO CNN \cite{Zimmermann2019FreiHANDAD} & CS & 11.0 & 10.9 & 71.3 & 71.5 \\ 
I2L-MeshNet \cite{Moon2020I2LMeshNetIP} & CS & 7.4 & 7.6 & 60.3 & 60.4 \\ 
CMR-PG \cite{Chen2021CameraSpaceHM} & CS & 6.9 & 7.0 & 48.9 & 49.0 \\ 
\hline
Ours(ResNet18) & CS & 7.4 & 7.5 & 49.2 & 49.3 \\
Ours(ResNet50) & CS & 6.9 & 7.2 & \textbf{47.1} & \textbf{47.1} \\
\hline
\end{tabular}}
\end{table*}

The PCK scores with thresholds from 0 to 50mm are shown in Figure~\ref{fig:pck}. In FreiHAND dataset, our approach surpasses all four existing methods in terms of the performance criteria in the camera space \cite{Hasson2019LearningJR, Zimmermann2019FreiHANDAD, Moon2020I2LMeshNetIP, Chen2021CameraSpaceHM} across all error thresholds within the dataset. Although our method demonstrates slightly lower performance compared to the best method, METRO \cite{Lin2020EndtoEndHP}, in root-relative recovery, the performance curves closely align. Furthermore, our approach outperforms the other five methods \cite{Hasson2019LearningJR, Zimmermann2019FreiHANDAD, Moon2020I2LMeshNetIP, Choi2020Pose2MeshGC, Chen2021CameraSpaceHM} in the same evaluation.

The comparison between our method and other methods on the RHD dataset is shown in Figure \ref{pck_rhd}. Employing the PA-MPJPE criterion, we analyze the predicted 3D hand joints and present the results within the 20-50mm error range for clarity. Our model achieved an AUC of 0.955, slightly lower than the method proposed by Kulon et al.\cite{Kulon2020WeaklySupervisedMH} but generally outperformed other approaches. The results validate the cross-domain generalization capability of our proposed model.


\subsection{Ablation Study}

\begin{table*}[!t]
\label{tab:ablation}
\caption{Ablation study. Overall ablation study numerical results. We evaluate the influence of 3 key components in out model along 6 different configurations on FreiHand Dataset.}
\centering
\scalebox{1.0}{\begin{tabular}{ c c c | c c c c}
\hline
Implicit Heatmap & Inception Graph & G\&L Attention & PJ & PV & CJ & CV \\
\hline
\ding{56} & \ding{56} & \ding{56} & 7.6 & 7.8 & 51.0 & 51.0 \\ 
\ding{56} & \checkmark & \ding{56} &  \textbf{7.4} & \textbf{7.5} & 50.6 & 50.6 \\ 
\ding{56} & \checkmark & \checkmark &  \textbf{7.4} & 7.6 & 49.8 & 49.9 \\ 
\checkmark & \ding{56} & \ding{56} &  7.8 & 7.9 & 49.9 & 50.0 \\ 
\checkmark & \checkmark & \ding{56} &  7.5 & 7.6 & 50.1 & 50.1 \\ 
\checkmark & \checkmark & \checkmark &  \textbf{7.4} & \textbf{7.5} & \textbf{49.2} & \textbf{49.3} \\ 
\hline
\end{tabular}}
\end{table*}

In order to validate the effectiveness of our proposed network framework, ablation experiments are carried out using the same conditions. The component of implicit heatmap, inception graph and G\&L attention in our model are evaluated. The performance of our models with different combinations of these components is shown in Table \ref{tab:ablation}. Due to the direct impact of root-relative results on the primary morphological representation of the hand, we chose not to include a separate comparison involving improvements in root depth prediction.

Implicit Heatmap: the first and fourth rows in Table \ref{tab:ablation} represent the results of our model with and without Implicit Heatmap. In the "w/o Implicit Heatmap" scenario, PJ and PV performed better with scores of 7.6 and 7.8 respectively. However, CJ and CV showed superior performance in the "w Implicit Heatmap" scenario, reducing errors by 1 compared to the "w/o Implicit Heatmap" scenario. By considering the third and sixth rows, PJ and PV of both models are similar, while CJ and CV show improved performance in the model using Implicit Heatmap, with scores of 49.2 and 49.3 respectively. These results suggest that directly incorporating the heatmap as part of relay supervision can provide some assistance in root-relative estimation. However, it is important to note that the gradient constraints generated by the heatmap can contaminate spatial information around the nodes to some extent. Consequently, this contamination affects spatial localization decisions and leads to a decrease in the accuracy of camera coordinate estimation.

Inception Graph: the models with aggregating k-neighborhoods using Inception graph convolution networks are indicated in the second and fifth rows, while the models using fixed k-order neighborhood convolution layers are shown in the first and fouth rows in Table \ref{tab:ablation}. The results illustrate that our proposed Inception Graph module has led to improvements in the performance of PJ and PV, with the best outcomes being 7.4 and 7.5 respectively. On average, this enhancement amounts to approximately 2mm in prediction accuracy. These findings highlight the effectiveness of our proposed Inception Graph Conv module, showcasing its ability to enhance the precision of root-relative hand predictions.

G\&L Attention: our proposed soft attention module in the deep reduction net is compared with a network that solely focuses on global information. In this comparison network, similar to Adabins, the input comprises 2D aggregated high-resolution cues. It utilizes point-wise dot-product attention operations between query and key vectors after convolutional encoding to acquire adaptive global information, while the bin width division is predicted by the query's head vector. By comparing the results indicated in the second and third rows, and the fifth and sixth rows of Table \ref{tab:ablation}, we can observe improvements in the scores for CJ and CV by using G\&L Attention. Specifically, CJ score improves from 50.6 to 49.8, while CV score improves from 50.6 to 49.9 when considering our proposed attention modulus. Similarly, the scores for CJ improve from 50.1 to 49.2, and for CV from 50.1 to 49.3. This performance gap clearly demonstrates the effectiveness of our feature attention interaction module in the depth estimation pipeline. Aggregating global and local features from the images and hand regions successfully enhances the performance of the model.


\subsection{Visualization}

\begin{figure*}[t]
\centering
\includegraphics[width=4.5in]{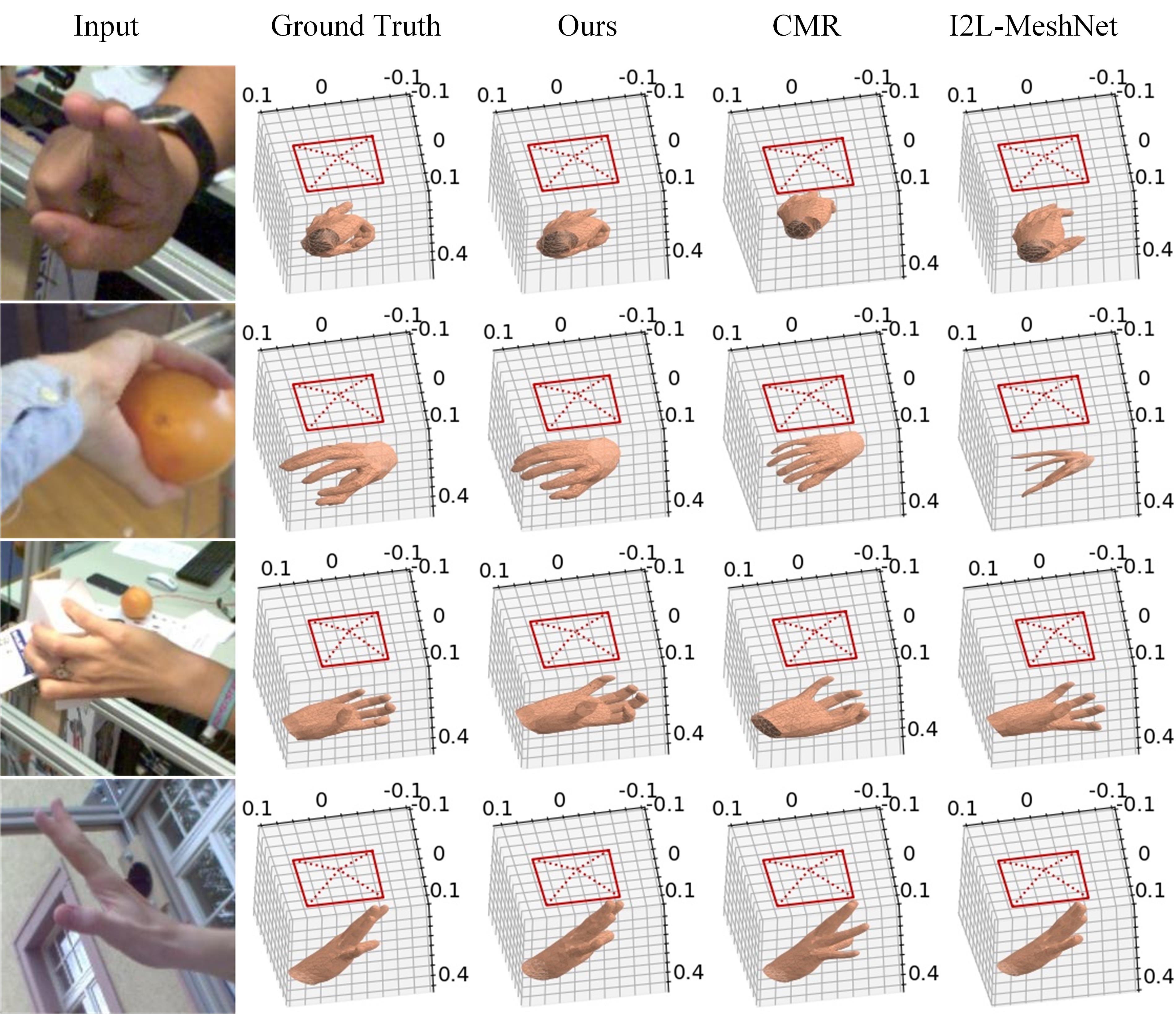}
\caption{Visualization of the recovered meshes.}
\label{fig_compare}
\end{figure*}

The 3D hand meshes recovered by different methods are compared visually in Figure \ref{fig_compare}. Four samples with complex backgrounds and self-occlusion are chosen for illustration. The columns indicate the original images, the ground truth, the recovered meshes of our method and two currently prominent methods (CMR and I2L-MeshNet). The red square represents the camera position, and the intersection point represents the camera space origin. The figures shows that our network yields superior results. The predicted shapes closely resemble the ground truth, exhibiting a high level of accuracy in spatial positioning. Consequently, the predicted shapes seamlessly blend into the environment. This comparison provides evidence that our model outperforms previously proposed models in scenarios involving complex backgrounds and self-occlusion. Our model demonstrates superior capabilities in reconstructing hand shapes and spatial localization under these challenging conditions.

\section{Conclusion}
\label{sec:conclusion}

This study proposes a network model that parallelizes the tasks of root-relative grid and root recovery, enabling the three-dimensional hand mesh recovery in camera coordinate space. To ensure compatibility and effective utilization of the collected 2D information across different subtasks, an implicit learning approach is adopted for 2D heatmaps. This approach strikes a balance between high-level semantic information and fine details. Additionally, a spectral graph convolutional network method inspired by the Inception concept is devised, which enhances the model's predictive performance in handling complex environments and self-occluded scenes. The depth estimation is formulated as a bin classification problem, and both local details and global feature are incorporated by our proposed attention mechanisms. The experimental results confirms that the exceptional performance of our model in both evaluations of root-relative and camera space evaluations and our model outperforms the SOTA methods of 3D hand mesh recovery. The approach presented in this work establishes a fundamental framework for predicting absolute spatial grid coordinates. This framework enables end-to-end prediction in two-stage methods, significantly reducing the complexity associated with algorithm deployment and management.

One of the possible future works is to improve our model by constructing lightweight deep neural networks. This optimization will facilitate seamless integration on mobile devices, addressing computational constraints and ensuring efficient real-time processing.


\backmatter

\bigskip


\bibliography{ref.bib}


\begin{thebibliography}{42}
\ifx \bisbn   \undefined \def \bisbn  #1{ISBN #1}\fi
\ifx \binits  \undefined \def \binits#1{#1}\fi
\ifx \bauthor  \undefined \def \bauthor#1{#1}\fi
\ifx \batitle  \undefined \def \batitle#1{#1}\fi
\ifx \bjtitle  \undefined \def \bjtitle#1{#1}\fi
\ifx \bvolume  \undefined \def \bvolume#1{\textbf{#1}}\fi
\ifx \byear  \undefined \def \byear#1{#1}\fi
\ifx \bissue  \undefined \def \bissue#1{#1}\fi
\ifx \bfpage  \undefined \def \bfpage#1{#1}\fi
\ifx \blpage  \undefined \def \blpage #1{#1}\fi
\ifx \burl  \undefined \def \burl#1{\textsf{#1}}\fi
\ifx \doiurl  \undefined \def \doiurl#1{\url{https://doi.org/#1}}\fi
\ifx \betal  \undefined \def \betal{\textit{et al.}}\fi
\ifx \binstitute  \undefined \def \binstitute#1{#1}\fi
\ifx \binstitutionaled  \undefined \def \binstitutionaled#1{#1}\fi
\ifx \bctitle  \undefined \def \bctitle#1{#1}\fi
\ifx \beditor  \undefined \def \beditor#1{#1}\fi
\ifx \bpublisher  \undefined \def \bpublisher#1{#1}\fi
\ifx \bbtitle  \undefined \def \bbtitle#1{#1}\fi
\ifx \bedition  \undefined \def \bedition#1{#1}\fi
\ifx \bseriesno  \undefined \def \bseriesno#1{#1}\fi
\ifx \blocation  \undefined \def \blocation#1{#1}\fi
\ifx \bsertitle  \undefined \def \bsertitle#1{#1}\fi
\ifx \bsnm \undefined \def \bsnm#1{#1}\fi
\ifx \bsuffix \undefined \def \bsuffix#1{#1}\fi
\ifx \bparticle \undefined \def \bparticle#1{#1}\fi
\ifx \barticle \undefined \def \barticle#1{#1}\fi
\bibcommenthead
\ifx \bconfdate \undefined \def \bconfdate #1{#1}\fi
\ifx \botherref \undefined \def \botherref #1{#1}\fi
\ifx \url \undefined \def \url#1{\textsf{#1}}\fi
\ifx \bchapter \undefined \def \bchapter#1{#1}\fi
\ifx \bbook \undefined \def \bbook#1{#1}\fi
\ifx \bcomment \undefined \def \bcomment#1{#1}\fi
\ifx \oauthor \undefined \def \oauthor#1{#1}\fi
\ifx \citeauthoryear \undefined \def \citeauthoryear#1{#1}\fi
\ifx \endbibitem  \undefined \def \endbibitem {}\fi
\ifx \bconflocation  \undefined \def \bconflocation#1{#1}\fi
\ifx \arxivurl  \undefined \def \arxivurl#1{\textsf{#1}}\fi
\csname PreBibitemsHook\endcsname

\bibitem[\protect\citeauthoryear{Kong et~al.}{2022}]{Kong2022IdentityAwareHM}
\begin{botherref}
\oauthor{\bsnm{Kong}, \binits{D.}},
\oauthor{\bsnm{Zhang}, \binits{L.}},
\oauthor{\bsnm{Chen}, \binits{L.}},
\oauthor{\bsnm{Ma}, \binits{H.}},
\oauthor{\bsnm{Yan}, \binits{X.}},
\oauthor{\bsnm{Sun}, \binits{S.}},
\oauthor{\bsnm{Liu}, \binits{X.}},
\oauthor{\bsnm{Han}, \binits{K.}},
\oauthor{\bsnm{Xie}, \binits{X.}}:
Identity-aware hand mesh estimation and personalization from rgb images.
ArXiv
\textbf{abs/2209.10840}
(2022)
\end{botherref}
\endbibitem

\bibitem[\protect\citeauthoryear{Xu et~al.}{2020}]{Xu2020BuildingHH}
\begin{barticle}
\bauthor{\bsnm{Xu}, \binits{Z.}},
\bauthor{\bsnm{Chang}, \binits{W.-F.H.J.-K.}},
\bauthor{\bsnm{Zhu}, \binits{Y.}},
\bauthor{\bsnm{Dong}, \binits{L.}},
\bauthor{\bsnm{Zhou}, \binits{H.}},
\bauthor{\bsnm{Zhang}, \binits{Q.}}:
\batitle{Building high-fidelity human body models from user-generated data}.
\bjtitle{IEEE Transactions on Multimedia}
\bvolume{23},
\bfpage{1542}--\blpage{1556}
(\byear{2020})
\end{barticle}
\endbibitem

\bibitem[\protect\citeauthoryear{Choi et~al.}{2020}]{Choi2020Pose2MeshGC}
\begin{botherref}
\oauthor{\bsnm{Choi}, \binits{H.}},
\oauthor{\bsnm{Moon}, \binits{G.}},
\oauthor{\bsnm{Lee}, \binits{K.M.}}:
Pose2mesh: Graph convolutional network for 3d human pose and mesh recovery from a 2d human pose.
ArXiv
\textbf{abs/2008.09047}
(2020)
\end{botherref}
\endbibitem

\bibitem[\protect\citeauthoryear{Ge et~al.}{2019}]{Ge20193DHS}
\begin{botherref}
\oauthor{\bsnm{Ge}, \binits{L.}},
\oauthor{\bsnm{Ren}, \binits{Z.}},
\oauthor{\bsnm{Li}, \binits{Y.}},
\oauthor{\bsnm{Xue}, \binits{Z.}},
\oauthor{\bsnm{Wang}, \binits{Y.}},
\oauthor{\bsnm{Cai}, \binits{J.}},
\oauthor{\bsnm{Yuan}, \binits{J.}}:
3d hand shape and pose estimation from a single rgb image.
2019 IEEE/CVF Conference on Computer Vision and Pattern Recognition (CVPR),
10825--10834
(2019)
\end{botherref}
\endbibitem

\bibitem[\protect\citeauthoryear{Boukhayma et~al.}{2019}]{Boukhayma20193DHS}
\begin{botherref}
\oauthor{\bsnm{Boukhayma}, \binits{A.}},
\oauthor{\bsnm{Bem}, \binits{R.}},
\oauthor{\bsnm{Torr}, \binits{P.H.S.}}:
3d hand shape and pose from images in the wild.
2019 IEEE/CVF Conference on Computer Vision and Pattern Recognition (CVPR),
10835--10844
(2019)
\end{botherref}
\endbibitem

\bibitem[\protect\citeauthoryear{Lin et~al.}{2018}]{Lin2018AFM}
\begin{botherref}
\oauthor{\bsnm{Lin}, \binits{C.}},
\oauthor{\bsnm{Andersen}, \binits{D.}},
\oauthor{\bsnm{Popescu}, \binits{V.}},
\oauthor{\bsnm{Rojas-Mu{\~n}oz}, \binits{E.}},
\oauthor{\bsnm{Cabrera}, \binits{M.E.}},
\oauthor{\bsnm{Mullis}, \binits{B.H.}},
\oauthor{\bsnm{Zarzaur}, \binits{B.}},
\oauthor{\bsnm{Anderson}, \binits{K.}},
\oauthor{\bsnm{Marley}, \binits{S.}},
\oauthor{\bsnm{Wachs}, \binits{J.P.}}:
A first-person mentee second-person mentor ar interface for surgical telementoring.
2018 IEEE International Symposium on Mixed and Augmented Reality Adjunct (ISMAR-Adjunct),
3--8
(2018)
\end{botherref}
\endbibitem

\bibitem[\protect\citeauthoryear{Moon and Lee}{2020}]{Moon2020I2LMeshNetIP}
\begin{botherref}
\oauthor{\bsnm{Moon}, \binits{G.}},
\oauthor{\bsnm{Lee}, \binits{K.M.}}:
I2l-meshnet: Image-to-lixel prediction network for accurate 3d human pose and mesh estimation from a single rgb image.
ArXiv
\textbf{abs/2008.03713}
(2020)
\end{botherref}
\endbibitem

\bibitem[\protect\citeauthoryear{Chen et~al.}{2021}]{Chen2021CameraSpaceHM}
\begin{botherref}
\oauthor{\bsnm{Chen}, \binits{X.}},
\oauthor{\bsnm{Liu}, \binits{Y.}},
\oauthor{\bsnm{Ma}, \binits{C.}},
\oauthor{\bsnm{Chang}, \binits{J.}},
\oauthor{\bsnm{Wang}, \binits{H.}},
\oauthor{\bsnm{Chen}, \binits{T.}},
\oauthor{\bsnm{Guo}, \binits{X.}},
\oauthor{\bsnm{Wan}, \binits{P.}},
\oauthor{\bsnm{Zheng}, \binits{W.}}:
Camera-space hand mesh recovery via semantic aggregation and adaptive 2d-1d registration.
2021 IEEE/CVF Conference on Computer Vision and Pattern Recognition (CVPR),
13269--13278
(2021)
\end{botherref}
\endbibitem

\bibitem[\protect\citeauthoryear{Iqbal et~al.}{2018}]{Iqbal2018HandPE}
\begin{bchapter}
\bauthor{\bsnm{Iqbal}, \binits{U.}},
\bauthor{\bsnm{Molchanov}, \binits{P.}},
\bauthor{\bsnm{Breuel}, \binits{T.M.}},
\bauthor{\bsnm{Gall}, \binits{J.}},
\bauthor{\bsnm{Kautz}, \binits{J.}}:
\bctitle{Hand pose estimation via latent 2.5d heatmap regression}.
In: \bbtitle{European Conference on Computer Vision}
(\byear{2018}).
\burl{https://api.semanticscholar.org/CorpusID:13746398}
\end{bchapter}
\endbibitem

\bibitem[\protect\citeauthoryear{Hasson et~al.}{2019}]{Hasson2019LearningJR}
\begin{botherref}
\oauthor{\bsnm{Hasson}, \binits{Y.}},
\oauthor{\bsnm{Varol}, \binits{G.}},
\oauthor{\bsnm{Tzionas}, \binits{D.}},
\oauthor{\bsnm{Kalevatykh}, \binits{I.}},
\oauthor{\bsnm{Black}, \binits{M.J.}},
\oauthor{\bsnm{Laptev}, \binits{I.}},
\oauthor{\bsnm{Schmid}, \binits{C.}}:
Learning joint reconstruction of hands and manipulated objects.
2019 IEEE/CVF Conference on Computer Vision and Pattern Recognition (CVPR),
11799--11808
(2019)
\end{botherref}
\endbibitem

\bibitem[\protect\citeauthoryear{Hasson et~al.}{2021}]{Hasson2021TowardsUJ}
\begin{botherref}
\oauthor{\bsnm{Hasson}, \binits{Y.}},
\oauthor{\bsnm{Varol}, \binits{G.}},
\oauthor{\bsnm{Laptev}, \binits{I.}},
\oauthor{\bsnm{Schmid}, \binits{C.}}:
Towards unconstrained joint hand-object reconstruction from rgb videos.
2021 International Conference on 3D Vision (3DV),
659--668
(2021)
\end{botherref}
\endbibitem

\bibitem[\protect\citeauthoryear{Moon et~al.}{2019}]{Moon2019CameraDT}
\begin{botherref}
\oauthor{\bsnm{Moon}, \binits{G.}},
\oauthor{\bsnm{Chang}, \binits{J.Y.}},
\oauthor{\bsnm{Lee}, \binits{K.M.}}:
Camera distance-aware top-down approach for 3d multi-person pose estimation from a single rgb image.
2019 IEEE/CVF International Conference on Computer Vision (ICCV),
10132--10141
(2019)
\end{botherref}
\endbibitem

\bibitem[\protect\citeauthoryear{Kim et~al.}{2022}]{Kim2022GlobalLocalPN}
\begin{botherref}
\oauthor{\bsnm{Kim}, \binits{D.}},
\oauthor{\bsnm{Ga}, \binits{W.-S.}},
\oauthor{\bsnm{Ahn}, \binits{P.}},
\oauthor{\bsnm{Joo}, \binits{D.}},
\oauthor{\bsnm{Chun}, \binits{S.Y.}},
\oauthor{\bsnm{Kim}, \binits{J.}}:
Global-local path networks for monocular depth estimation with vertical cutdepth.
ArXiv
\textbf{abs/2201.07436}
(2022)
\end{botherref}
\endbibitem

\bibitem[\protect\citeauthoryear{Xu and Takano}{2021}]{Xu2021GraphSH}
\begin{botherref}
\oauthor{\bsnm{Xu}, \binits{T.}},
\oauthor{\bsnm{Takano}, \binits{W.}}:
Graph stacked hourglass networks for 3d human pose estimation.
2021 IEEE/CVF Conference on Computer Vision and Pattern Recognition (CVPR),
16100--16109
(2021)
\end{botherref}
\endbibitem

\bibitem[\protect\citeauthoryear{Romero et~al.}{2017}]{Romero2017EmbodiedH}
\begin{barticle}
\bauthor{\bsnm{Romero}, \binits{J.}},
\bauthor{\bsnm{Tzionas}, \binits{D.}},
\bauthor{\bsnm{Black}, \binits{M.J.}}:
\batitle{Embodied hands}.
\bjtitle{ACM Transactions on Graphics (TOG)}
\bvolume{36},
\bfpage{1}--\blpage{17}
(\byear{2017})
\end{barticle}
\endbibitem

\bibitem[\protect\citeauthoryear{Defferrard et~al.}{2016}]{Defferrard2016ConvolutionalNN}
\begin{bchapter}
\bauthor{\bsnm{Defferrard}, \binits{M.}},
\bauthor{\bsnm{Bresson}, \binits{X.}},
\bauthor{\bsnm{Vandergheynst}, \binits{P.}}:
\bctitle{Convolutional neural networks on graphs with fast localized spectral filtering}.
In: \bbtitle{NIPS}
(\byear{2016}).
\burl{https://api.semanticscholar.org/CorpusID:3016223}
\end{bchapter}
\endbibitem

\bibitem[\protect\citeauthoryear{Vaswani et~al.}{2017}]{Vaswani2017AttentionIA}
\begin{bchapter}
\bauthor{\bsnm{Vaswani}, \binits{A.}},
\bauthor{\bsnm{Shazeer}, \binits{N.M.}},
\bauthor{\bsnm{Parmar}, \binits{N.}},
\bauthor{\bsnm{Uszkoreit}, \binits{J.}},
\bauthor{\bsnm{Jones}, \binits{L.}},
\bauthor{\bsnm{Gomez}, \binits{A.N.}},
\bauthor{\bsnm{Kaiser}, \binits{L.}},
\bauthor{\bsnm{Polosukhin}, \binits{I.}}:
\bctitle{Attention is all you need}.
In: \bbtitle{NIPS}
(\byear{2017}).
\burl{https://api.semanticscholar.org/CorpusID:13756489}
\end{bchapter}
\endbibitem

\bibitem[\protect\citeauthoryear{Li et~al.}{2022}]{li2022detailed}
\begin{barticle}
\bauthor{\bsnm{Li}, \binits{Z.}},
\bauthor{\bsnm{Oskarsson}, \binits{M.}},
\bauthor{\bsnm{Heyden}, \binits{A.}}:
\batitle{Detailed 3d human body reconstruction from multi-view images combining voxel super-resolution and learned implicit representation}.
\bjtitle{Applied Intelligence}
\bvolume{52}(\bissue{6}),
\bfpage{6739}--\blpage{6759}
(\byear{2022})
\end{barticle}
\endbibitem

\bibitem[\protect\citeauthoryear{Zhou et~al.}{2020}]{Zhou2020MonocularRH}
\begin{botherref}
\oauthor{\bsnm{Zhou}, \binits{Y.}},
\oauthor{\bsnm{Habermann}, \binits{M.}},
\oauthor{\bsnm{Xu}, \binits{W.}},
\oauthor{\bsnm{Habibie}, \binits{I.}},
\oauthor{\bsnm{Theobalt}, \binits{C.}},
\oauthor{\bsnm{Xu}, \binits{F.}}:
Monocular real-time hand shape and motion capture using multi-modal data.
2020 IEEE/CVF Conference on Computer Vision and Pattern Recognition (CVPR),
5345--5354
(2020)
\end{botherref}
\endbibitem

\bibitem[\protect\citeauthoryear{Chen and Sun}{2023}]{chen2023joint}
\begin{barticle}
\bauthor{\bsnm{Chen}, \binits{Z.}},
\bauthor{\bsnm{Sun}, \binits{Y.}}:
\batitle{Joint-wise 2d to 3d lifting for hand pose estimation from a single rgb image}.
\bjtitle{Applied Intelligence}
\bvolume{53}(\bissue{6}),
\bfpage{6421}--\blpage{6431}
(\byear{2023})
\end{barticle}
\endbibitem

\bibitem[\protect\citeauthoryear{Zhang and Zhang}{2019}]{Zhang2019PixelwiseR3}
\begin{botherref}
\oauthor{\bsnm{Zhang}, \binits{X.}},
\oauthor{\bsnm{Zhang}, \binits{F.}}:
Pixel-wise regression: 3d hand pose estimation via spatial-form representation and differentiable decoder.
ArXiv
\textbf{abs/1905.02085}
(2019)
\end{botherref}
\endbibitem

\bibitem[\protect\citeauthoryear{Lim et~al.}{2018}]{Lim2018ASA}
\begin{botherref}
\oauthor{\bsnm{Lim}, \binits{I.}},
\oauthor{\bsnm{Dielen}, \binits{A.C.A.}},
\oauthor{\bsnm{Campen}, \binits{M.}},
\oauthor{\bsnm{Kobbelt}, \binits{L.}}:
A simple approach to intrinsic correspondence learning on unstructured 3d meshes.
ArXiv
\textbf{abs/1809.06664}
(2018)
\end{botherref}
\endbibitem

\bibitem[\protect\citeauthoryear{Kourbane and Genc}{2022}]{kourbane2022graph}
\begin{barticle}
\bauthor{\bsnm{Kourbane}, \binits{I.}},
\bauthor{\bsnm{Genc}, \binits{Y.}}:
\batitle{A graph-based approach for absolute 3d hand pose estimation using a single rgb image}.
\bjtitle{Applied Intelligence}
\bvolume{52}(\bissue{14}),
\bfpage{16667}--\blpage{16682}
(\byear{2022})
\end{barticle}
\endbibitem

\bibitem[\protect\citeauthoryear{Huang et~al.}{2023}]{Huang2023NeuralVF}
\begin{botherref}
\oauthor{\bsnm{Huang}, \binits{L.}},
\oauthor{\bsnm{Lin}, \binits{C.-C.}},
\oauthor{\bsnm{Lin}, \binits{K.}},
\oauthor{\bsnm{Liang}, \binits{L.}},
\oauthor{\bsnm{Wang}, \binits{L.}},
\oauthor{\bsnm{Yuan}, \binits{J.}},
\oauthor{\bsnm{Liu}, \binits{Z.}}:
Neural voting field for camera-space 3d hand pose estimation.
ArXiv
\textbf{abs/2305.04328}
(2023)
\end{botherref}
\endbibitem

\bibitem[\protect\citeauthoryear{Fu et~al.}{2018}]{Fu2018DeepOR}
\begin{botherref}
\oauthor{\bsnm{Fu}, \binits{H.}},
\oauthor{\bsnm{Gong}, \binits{M.}},
\oauthor{\bsnm{Wang}, \binits{C.}},
\oauthor{\bsnm{Batmanghelich}, \binits{K.}},
\oauthor{\bsnm{Tao}, \binits{D.}}:
Deep ordinal regression network for monocular depth estimation.
2018 IEEE/CVF Conference on Computer Vision and Pattern Recognition,
2002--2011
(2018)
\end{botherref}
\endbibitem

\bibitem[\protect\citeauthoryear{Lin and Lee}{2020}]{Lin2020HDNetHD}
\begin{bchapter}
\bauthor{\bsnm{Lin}, \binits{J.}},
\bauthor{\bsnm{Lee}, \binits{G.H.}}:
\bctitle{Hdnet: Human depth estimation for multi-person camera-space localization}.
In: \bbtitle{European Conference on Computer Vision}
(\byear{2020}).
\burl{https://api.semanticscholar.org/CorpusID:220633382}
\end{bchapter}
\endbibitem

\bibitem[\protect\citeauthoryear{Bhat et~al.}{2020}]{Bhat2020AdaBinsDE}
\begin{botherref}
\oauthor{\bsnm{Bhat}, \binits{S.}},
\oauthor{\bsnm{Alhashim}, \binits{I.}},
\oauthor{\bsnm{Wonka}, \binits{P.}}:
Adabins: Depth estimation using adaptive bins.
2021 IEEE/CVF Conference on Computer Vision and Pattern Recognition (CVPR),
4008--4017
(2020)
\end{botherref}
\endbibitem

\bibitem[\protect\citeauthoryear{He et~al.}{2015}]{He2015DeepRL}
\begin{botherref}
\oauthor{\bsnm{He}, \binits{K.}},
\oauthor{\bsnm{Zhang}, \binits{X.}},
\oauthor{\bsnm{Ren}, \binits{S.}},
\oauthor{\bsnm{Sun}, \binits{J.}}:
Deep residual learning for image recognition.
2016 IEEE Conference on Computer Vision and Pattern Recognition (CVPR),
770--778
(2015)
\end{botherref}
\endbibitem

\bibitem[\protect\citeauthoryear{Chen et~al.}{2021}]{Chen2021MobReconMH}
\begin{botherref}
\oauthor{\bsnm{Chen}, \binits{X.}},
\oauthor{\bsnm{Liu}, \binits{Y.}},
\oauthor{\bsnm{Dong}, \binits{Y.}},
\oauthor{\bsnm{Zhang}, \binits{X.}},
\oauthor{\bsnm{Ma}, \binits{C.}},
\oauthor{\bsnm{Xiong}, \binits{Y.}},
\oauthor{\bsnm{Zhang}, \binits{Y.}},
\oauthor{\bsnm{Guo}, \binits{X.}}:
Mobrecon: Mobile-friendly hand mesh reconstruction from monocular image.
2022 IEEE/CVF Conference on Computer Vision and Pattern Recognition (CVPR),
20512--20522
(2021)
\end{botherref}
\endbibitem

\bibitem[\protect\citeauthoryear{Nibali et~al.}{2018}]{Nibali2018NumericalCR}
\begin{botherref}
\oauthor{\bsnm{Nibali}, \binits{A.}},
\oauthor{\bsnm{He}, \binits{Z.}},
\oauthor{\bsnm{Morgan}, \binits{S.}},
\oauthor{\bsnm{Prendergast}, \binits{L.A.}}:
Numerical coordinate regression with convolutional neural networks.
ArXiv
\textbf{abs/1801.07372}
(2018)
\end{botherref}
\endbibitem

\bibitem[\protect\citeauthoryear{Newell et~al.}{2016}]{Newell2016StackedHN}
\begin{bchapter}
\bauthor{\bsnm{Newell}, \binits{A.}},
\bauthor{\bsnm{Yang}, \binits{K.}},
\bauthor{\bsnm{Deng}, \binits{J.}}:
\bctitle{Stacked hourglass networks for human pose estimation}.
In: \bbtitle{European Conference on Computer Vision}
(\byear{2016}).
\burl{https://api.semanticscholar.org/CorpusID:13613792}
\end{bchapter}
\endbibitem

\bibitem[\protect\citeauthoryear{Toshev and Szegedy}{2013}]{Toshev2013DeepPoseHP}
\begin{botherref}
\oauthor{\bsnm{Toshev}, \binits{A.}},
\oauthor{\bsnm{Szegedy}, \binits{C.}}:
Deeppose: Human pose estimation via deep neural networks.
2014 IEEE Conference on Computer Vision and Pattern Recognition,
1653--1660
(2013)
\end{botherref}
\endbibitem

\bibitem[\protect\citeauthoryear{Zhang et~al.}{2021}]{Zhang2021InteractingT3}
\begin{botherref}
\oauthor{\bsnm{Zhang}, \binits{B.}},
\oauthor{\bsnm{Wang}, \binits{Y.}},
\oauthor{\bsnm{Deng}, \binits{X.}},
\oauthor{\bsnm{Zhang}, \binits{Y.}},
\oauthor{\bsnm{Tan}, \binits{P.}},
\oauthor{\bsnm{Ma}, \binits{C.}},
\oauthor{\bsnm{Wang}, \binits{H.}}:
Interacting two-hand 3d pose and shape reconstruction from single color image.
2021 IEEE/CVF International Conference on Computer Vision (ICCV),
11334--11343
(2021)
\end{botherref}
\endbibitem

\bibitem[\protect\citeauthoryear{Dhillon et~al.}{2007}]{Dhillon2007WeightedGC}
\begin{botherref}
\oauthor{\bsnm{Dhillon}, \binits{I.S.}},
\oauthor{\bsnm{Guan}, \binits{Y.}},
\oauthor{\bsnm{Kulis}, \binits{B.}}:
Weighted graph cuts without eigenvectors a multilevel approach.
IEEE Transactions on Pattern Analysis and Machine Intelligence
\textbf{29}
(2007)
\end{botherref}
\endbibitem

\bibitem[\protect\citeauthoryear{Hu et~al.}{2017}]{Hu2017SqueezeandExcitationN}
\begin{botherref}
\oauthor{\bsnm{Hu}, \binits{J.}},
\oauthor{\bsnm{Shen}, \binits{L.}},
\oauthor{\bsnm{Albanie}, \binits{S.}},
\oauthor{\bsnm{Sun}, \binits{G.}},
\oauthor{\bsnm{Wu}, \binits{E.}}:
Squeeze-and-excitation networks.
2018 IEEE/CVF Conference on Computer Vision and Pattern Recognition,
7132--7141
(2017)
\end{botherref}
\endbibitem

\bibitem[\protect\citeauthoryear{Lee et~al.}{2019}]{Lee2019FromBT}
\begin{botherref}
\oauthor{\bsnm{Lee}, \binits{J.H.}},
\oauthor{\bsnm{Han}, \binits{M.-K.}},
\oauthor{\bsnm{Ko}, \binits{D.W.}},
\oauthor{\bsnm{Suh}, \binits{I.H.}}:
From big to small: Multi-scale local planar guidance for monocular depth estimation.
ArXiv
\textbf{abs/1907.10326}
(2019)
\end{botherref}
\endbibitem

\bibitem[\protect\citeauthoryear{Fan et~al.}{2016}]{Fan2016APS}
\begin{botherref}
\oauthor{\bsnm{Fan}, \binits{H.}},
\oauthor{\bsnm{Su}, \binits{H.}},
\oauthor{\bsnm{Guibas}, \binits{L.J.}}:
A point set generation network for 3d object reconstruction from a single image.
2017 IEEE Conference on Computer Vision and Pattern Recognition (CVPR),
2463--2471
(2016)
\end{botherref}
\endbibitem

\bibitem[\protect\citeauthoryear{Zimmermann et~al.}{2019}]{Zimmermann2019FreiHANDAD}
\begin{botherref}
\oauthor{\bsnm{Zimmermann}, \binits{C.}},
\oauthor{\bsnm{Ceylan}, \binits{D.}},
\oauthor{\bsnm{Yang}, \binits{J.}},
\oauthor{\bsnm{Russell}, \binits{B.C.}},
\oauthor{\bsnm{Argus}, \binits{M.}},
\oauthor{\bsnm{Brox}, \binits{T.}}:
Freihand: A dataset for markerless capture of hand pose and shape from single rgb images.
2019 IEEE/CVF International Conference on Computer Vision (ICCV),
813--822
(2019)
\end{botherref}
\endbibitem

\bibitem[\protect\citeauthoryear{Zimmermann and Brox}{2017}]{Zimmermann2017LearningTE}
\begin{botherref}
\oauthor{\bsnm{Zimmermann}, \binits{C.}},
\oauthor{\bsnm{Brox}, \binits{T.}}:
Learning to estimate 3d hand pose from single rgb images.
2017 IEEE International Conference on Computer Vision (ICCV),
4913--4921
(2017)
\end{botherref}
\endbibitem

\bibitem[\protect\citeauthoryear{Kingma and Ba}{2014}]{Kingma2014AdamAM}
\begin{botherref}
\oauthor{\bsnm{Kingma}, \binits{D.P.}},
\oauthor{\bsnm{Ba}, \binits{J.}}:
Adam: A method for stochastic optimization.
CoRR
\textbf{abs/1412.6980}
(2014)
\end{botherref}
\endbibitem

\bibitem[\protect\citeauthoryear{Lin et~al.}{2020}]{Lin2020EndtoEndHP}
\begin{botherref}
\oauthor{\bsnm{Lin}, \binits{K.}},
\oauthor{\bsnm{Wang}, \binits{L.}},
\oauthor{\bsnm{Liu}, \binits{Z.}}:
End-to-end human pose and mesh reconstruction with transformers.
2021 IEEE/CVF Conference on Computer Vision and Pattern Recognition (CVPR),
1954--1963
(2020)
\end{botherref}
\endbibitem

\bibitem[\protect\citeauthoryear{Kulon et~al.}{2020}]{Kulon2020WeaklySupervisedMH}
\begin{botherref}
\oauthor{\bsnm{Kulon}, \binits{D.}},
\oauthor{\bsnm{G{\"u}ler}, \binits{R.A.}},
\oauthor{\bsnm{Kokkinos}, \binits{I.}},
\oauthor{\bsnm{Bronstein}, \binits{M.M.}},
\oauthor{\bsnm{Zafeiriou}, \binits{S.}}:
Weakly-supervised mesh-convolutional hand reconstruction in the wild.
2020 IEEE/CVF Conference on Computer Vision and Pattern Recognition (CVPR),
4989--4999
(2020)
\end{botherref}
\endbibitem

\end{thebibliography}
\vfill

\end{document}